\documentclass[letterpaper]{article} % DO NOT CHANGE THIS
\usepackage{aaai24}  % DO NOT CHANGE THIS
\usepackage{times}  % DO NOT CHANGE THIS
\usepackage{helvet}  % DO NOT CHANGE THIS
\usepackage{courier}  % DO NOT CHANGE THIS
\usepackage[hyphens]{url}  % DO NOT CHANGE THIS
\usepackage{graphicx} % DO NOT CHANGE THIS
\usepackage{natbib}  % DO NOT CHANGE THIS AND DO NOT ADD ANY OPTIONS TO IT
\usepackage{caption} % DO NOT CHANGE THIS AND DO NOT ADD ANY OPTIONS TO IT
\usepackage{algorithm}
\usepackage{algorithmic}
\usepackage{newfloat}
\usepackage{listings}
\DeclareCaptionStyle{ruled}{labelfont=normalfont,labelsep=colon,strut=off} % DO NOT CHANGE THIS
\floatstyle{ruled}
\newfloat{listing}{tb}{lst}{}
\floatname{listing}{Listing}
\usepackage{amsmath,amsfonts}
\usepackage{subfigure}

\usepackage{booktabs}       % professional-quality 

\usepackage{pifont}

\usepackage{multirow}
\usepackage{hyperref}

\nocopyright

\title{Preparing Lessons for Progressive Training on Language Models}
\author{
    Yu Pan\textsuperscript{\rm 1}\equalcontrib,
    Ye Yuan\textsuperscript{\rm 3,}\textsuperscript{\rm 4}\equalcontrib,
    Yichun Yin\textsuperscript{\rm 5},
    Jiaxin Shi\textsuperscript{\rm 6},
    Zenglin Xu\textsuperscript{\rm 1,}\textsuperscript{\rm 2}\equalcorresponding,\\
    Ming Zhang\textsuperscript{\rm 3,}\textsuperscript{\rm 4}\equalcorresponding,
    Lifeng Shang\textsuperscript{\rm 5},
    Xin Jiang\textsuperscript{\rm 5},
    Qun Liu\textsuperscript{\rm 5}
}
\affiliations{
    \textsuperscript{\rm 1}Harbin Institute of Technology Shenzhen, Shenzhen, Guangdong, China\\
    \textsuperscript{\rm 2}Pengcheng Laboratory, Shenzhen, China\\
    \textsuperscript{\rm 3}School of Computer Science, Peking University, Beijing, China\\
    \textsuperscript{\rm 4}Peking University-Anker Embodied AI Lab\\
    \textsuperscript{\rm 5}Huawei Noah’s Ark Lab, Shenzhen, Guangdong, China\\
    \textsuperscript{\rm 6}Cloud BU, Huawei Technologies
}

\usepackage{bibentry}
\def\d{\,\mathrm{d}}

\begin{document}

\maketitle

\begin{abstract}
The rapid progress of Transformers in artificial intelligence has come at the cost of increased resource consumption and greenhouse gas emissions due to growing model sizes. Prior work suggests using pretrained small models to improve training efficiency, but this approach may not be suitable for new model structures. On the other hand, training from scratch can be slow, and progressively stacking layers often fails to achieve significant acceleration. To address these challenges, we propose a novel method called Apollo, which prep\textbf{a}res lessons for ex\textbf{p}anding \textbf{o}perations by \textbf{l}earning high-\textbf{l}ayer functi\textbf{o}nality during training of low layers. Our approach involves low-value-prioritized sampling (LVPS) to train different depths and weight sharing to facilitate efficient expansion. We also introduce an interpolation method for stable model depth extension. Experiments demonstrate that Apollo achieves state-of-the-art acceleration ratios, even rivaling methods using pretrained models, making it a universal and efficient solution for training deep models while reducing time, financial, and environmental costs. Our source code is available in \url{https://github.com/yuanyehome/Apollo-AAAI-2024-Release}.
\end{abstract}

\section{Introduction}
\label{sec:intro}

Transformers~\citep{DBLP:conf/nips/VaswaniSPUJGKP17} have recently achieved a significant impact on the field of artificial intelligence~\citep{DBLP:conf/iconip/WangSL0ZX20,DBLP:journals/corr/abs-2009-10580,DBLP:conf/iclr/DosovitskiyB0WZ21,DBLP:journals/corr/abs-2105-05537,DBLP:journals/corr/abs-2302-09019}. Nevertheless, the training cost is increasing in terms of the growing model size, which causes an amount of resourcing consumption and greenhouse gases emission~\citep{DBLP:journals/corr/abs-1907-10597,DBLP:conf/aaai/PanXWYWBX19}. Addressing this problem, recent  work~\citep{DBLP:conf/acl/ChenYS0QWWCL022,DBLP:journals/corr/ChenGS15,wanglearning,DBLP:journals/corr/abs-2310-10699} suggests to improve training efficiency by reusing a  pretrained small model as an initialization method that contains knowledge prior. However, the requirement of a pretrained model can cause fatal obstacles, especially for a new-designed model structure, which affects the applicability of these studies as general strategies for training. On the other hand, the studies of training from scratch~\citep{DBLP:conf/icml/GongHLQWL19,DBLP:journals/corr/abs-2011-13635} generally stack layers to train Transformer progressively. Nevertheless, these methods usually cannot achieve significant acceleration in training, and are thus slower than training from pretrained models.  Against this background, it is an emergency to design a universal method to efficiently train the models with reduced time and financial costs, while benefiting the ecological environment.

To achieve this goal, the progressive expansion of models in depth proves to be a crucial aspect in training from scratch. One noteworthy example is StackBERT~\citep{DBLP:conf/icml/GongHLQWL19}, where a stacking learning strategy contributes to improved training efficiency through two merits: (1) Fewer layers in the initial stages of training require fewer computational resources, leading to faster training; (2) Lower trained weights provide a usable prior that benefits the training of stacked higher weights.
While StackBERT undeniably accelerates training from the second merit, there are still two concerns that need to be addressed. Firstly, the suitability of the stacking method is questionable. For instance, directly stacking the 1-st layer onto the 7-th layer of a 12-layer Transformer is not intuitive due to the clear differences in semantic functionality between them~\cite{DBLP:journals/tacl/RogersKR20}. Moreover, even though there might be some similarities across the entire model, it has been pointed out that most of the layers are different from each other~\citep{DBLP:conf/acl/ChenYS0QWWCL022}. Consequently, it raises doubts about whether the normally trained weights are sufficiently prepared well to be expanded effectively, given the lack of knowledge in higher layers.

\begin{figure*}[t]
\centering
\includegraphics[width=0.91\textwidth]{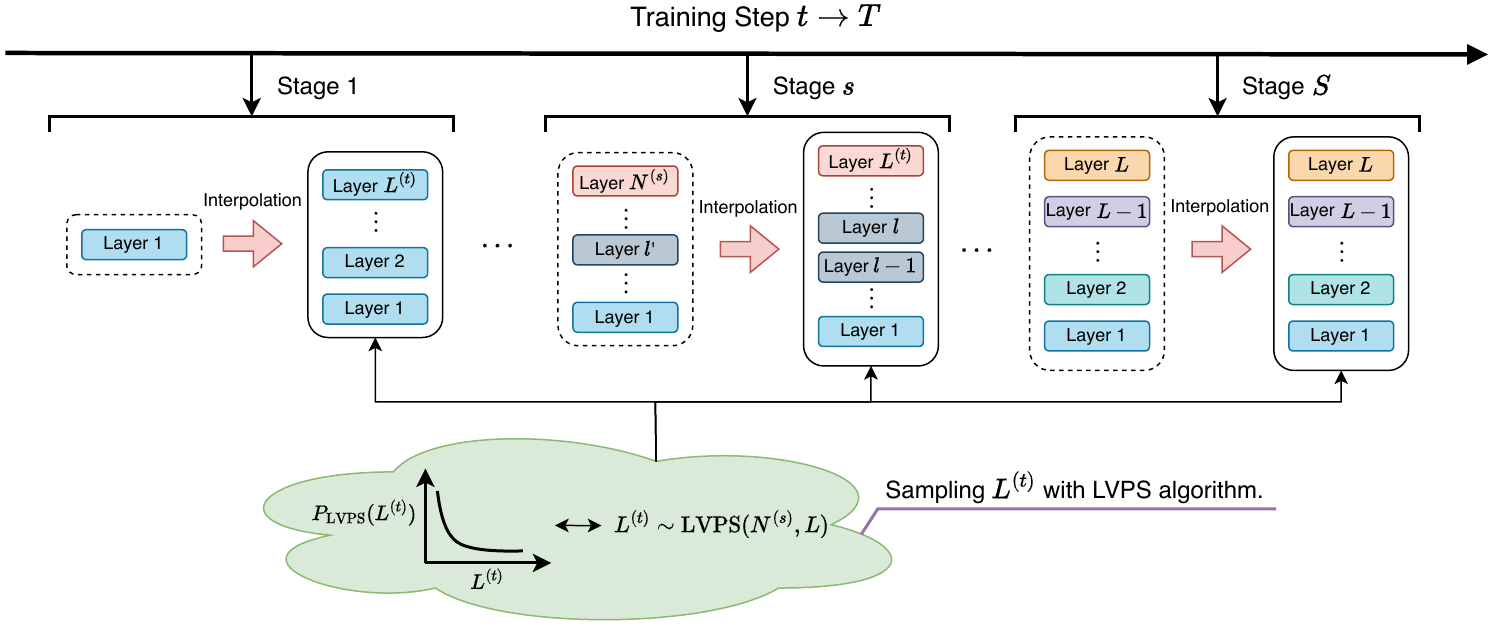}

\caption{An illustration of the Apollo for training an $L$-layered model within $T$ steps. We divide this training process into $S$ stages. In the $t$-th step at the $s$-th stage, the model weights are $N^{(s)}$ layers (the left layers in each stage in the figure). To let the $N^{(s)}$ layers learn functionality in high layers in advance, we construct $L^{(t)}$ layers (the right layers in each stage in the figure) by sharing the $N^{(s)}$ weights through an interpolation method, where $N^{(s)} \leq L^{(t)}$. As shown in the figure, the same color denotes the same weight. We randomly choose $L^{(t)}$ at $t$-th step through a probability function Low-Value-Prioritized Sampling (LVPS). Since LVPS tends to select shallower layers, it can greatly save computation costs. Furthermore, we progressively increase the $N^{(s)}$ weights when stepping into the next stage. Since weights in the early stage can learn the properties of higher layers, Apollo can significantly contribute to the training efficiency.
}
\label{fig:overview}
\end{figure*}

Motivated by this consideration, we introduce ``Apollo'' - a novel approach to preparing lessons for low-layer weights to learn the high-layer functionality in the training process. This strategy proves beneficial in further extending the capabilities of the model. In essence, Apollo involves two key components. Firstly, we employ a low-value-prioritized sampling (LVPS) technique, which randomly selects a depth for training at each step. This helps to ensure a diverse training experience. Subsequently, we share the low-layer weights, enabling them to adapt to the selected layers by LVPS. These shared weights are well-prepared not only for learning high-layer functionality but also for recurrent transformation, as supported by previous work~\cite{DBLP:conf/iclr/LanCGGSS20,DBLP:conf/iclr/DehghaniGVUK19,DBLP:journals/corr/abs-2110-03848}. It is worth noting that while weight-sharing strategies have been explored previously to facilitate information exchange across layers, our approach is novel in its dynamic application to sample layers, which is important to learn high-layer property to improve training efficiency. Furthermore, we address the issue of training stability by introducing an interpolation method to extend the depth of the model. This is essential since directly stacking layers can cause large gradients, which can be detrimental to training stability. In all, we summarize our contribution as:
\begin{itemize}
    \item Through sharing weights in the early stage to learn the functionality of high layers, Apollo effectively expands the depth of networks, resulting in remarkable training acceleration.
    \item Through LVPS, Apollo achieves a substantial reduction in training FLOPs by predominantly sampling low depth layers, while retaining the benefit of expanding depth.
    \item Through replacing layer stacking with layer interpolation, Apollo further enhances the stability of the expanded model.
    \item Experiments show that Apollo attains state-of-the-art training efficiency, surpassing even the methods reliant on pretrained models.
\end{itemize}

\section{Related Work}
\label{sec:related}

\paragraph{Efficient training from scratches.} 

Training from scratch means training without any prior knowledge. Some approaches~\citep{DBLP:conf/icml/GongHLQWL19,DBLP:journals/corr/abs-2011-13635,DBLP:conf/cvpr/LiZWLCY22,DBLP:conf/naacl/GuLYLCH21,DBLP:conf/icml/ShenWKDPB22} are known as ``progressive training,'' which involves initially pretraining a smaller scratch model and then gradually increasing its size, resulting in accelerated training. There are various training strategies for such models that are universally applicable and independent of our own method. For instance, employing optimization techniques like Adam~\citep{DBLP:journals/corr/KingmaB14} can accelerate the learning process by considering the optimizer's perspective. \citet{DBLP:journals/corr/abs-1909-08053} have successfully utilized mixed precision training to enhance training efficiency, while low-rank methods provide a viable option for memory and time-efficient training~\citep{DBLP:journals/corr/abs-2209-13569}. Additionally, \citet{DBLP:conf/iclr/0001XLKHL21} have demonstrated improved data efficiency by taking note of rare words. Other effective training methods include dropping layers~\citep{DBLP:conf/nips/ZhangH20}, knowledge inheritance~\citep{DBLP:conf/naacl/QinLYZ0ZSLLS022}, and merging tokens~\citep{DBLP:journals/corr/abs-2210-09461}. Our method embraces the concept of progressive training and attains noteworthy training efficiency by preparing instructive lessons to facilitate the expansion of layer depth.

\paragraph{Efficient training by reusing pretrained models.} 

Recent studies have demonstrated the great potentials of developing large pre-trained models via expanding a small  model. 
The pioneering work of Net2Net~\citep{DBLP:journals/corr/ChenGS15} introduced the concept of function-preserving transformations which increase the width via neuron splitting and the depth via identity layers. However, the randomly chosen neurons for splitting in Net2Net are not promised to be well performed. Addressing this challenge, a series of studies~\citep{DBLP:conf/nips/WuW019,DBLP:journals/corr/abs-2003-10392,DBLP:journals/corr/abs-1910-03103,DBLP:conf/nips/WuLS020} have adopted functional steepest descent to select the optimal subset of neurons for splitting. Following the idea of function preservation, bert2BERT~\citep{DBLP:conf/acl/ChenYS0QWWCL022} expanded small transformers. More recently, LiGO~\citep{wanglearning} introduced a trainable linear operator to learn an effective expansion formula. Mango~\citep{DBLP:journals/corr/abs-2310-10699} utilized a tensor ring matrix product operator (TR-MPO) to grow a small pretrained model to a large counterpart for efficient training. In the past, the utilization of pretrained methods consistently yielded greater acceleration than that achieved by training from scratch. Remarkably, our approach demonstrates highly promising results, surpassing even the swiftness of training from a pretrained model.

\section{Method}
\label{sec:method}

In this section, we will introduce our method to sample the information of the higher layer to accelerate training.

\subsection{Notations}
To describe the implementation of Apollo, we introduce related notations here. We denote an $L$-layered network by $f^{(L)}(\cdot)$, and denote the function of the $l$-th layer of $f^{(L)}$ as $f_l(\cdot)$, where $l\in [L]$. Thus, given an input $x$, $f^{(L)}(x)$ can be formulated as
\begin{align}
    f^{(L)}(x) = f_L(f_{L-1}(\dots f_l(\dots f_1(x)))).
\end{align}
We denote the set of $N$ weights in $f$ as $\{\theta_i\}^{N}_{i=1}$, and the weights of $l$-the layer as $\Theta(f_l)$. As our method samples information of higher layers through weight sharing, here we formally consider a mapping function $g(\cdot)$ to arrange the $g(l)$-th weight to the $l$-th layer as
\begin{align}
    \Theta(f_l) = \theta_{g(l)},
\end{align}
where $g(\cdot) \in [N]$. Moreover, for the convenience of expressing the process of layer expansion, we denote the layer size of the $t$-th step as $L^{(t)}$, where $t\in T$ and $T$ is the total steps. Thus, a network at the $t$-th step can be denoted by  $f^{(L^{(t)})}$.

\textbf{Transformer Architecture.} Here, we introduce the structure of the Transformer~\citep{DBLP:conf/nips/VaswaniSPUJGKP17}, each block of which mainly contains two basic layers, (1) the multi-head self-attention (MHSA) layer, (2) the feed-forward neural network (FFN) layer. Assuming inputs of $l$-th layer are query $\mathbf{Q}_l \in \mathbb{R}^{d}$, key $\mathbf{K}_l \in \mathbb{R}^{d}$, and value $\mathbf{V}_l \in \mathbb{R}^{d}$, where $d$ is the dimension, the $l$-th layer can be formulated as
\begin{align}
    f_l(\textbf{Q}_l, \textbf{K}_l, \textbf{V}_l) = \operatorname{FFN}(\operatorname{MHSA}(\textbf{Q}_l, \textbf{K}_l, \textbf{V}_l)).
\end{align}
The MHSA and FFN layers are defined as follows:

(1) MHSA Layer. The MHSA layer is another form of the Self-Attention (SA) layer that allows the model to weigh the importance of different words in a sentence, enabling the ability to capture long-range dependencies and contextual information effectively. Instead of using a single attention mechanism, MHSA divides SA into a multi-head structure. Each head attends to different aspects or dependencies in the input sequence. This allows the model to capture different patterns and relationships between words, potentially improving the model's ability to learn complex dependencies and relationships. Specifically, the parameters of SA in $l$-the layer are $\mathbf{W}_l^{\{Q, K, V, O\}} \in \mathbb{R}^{d\times d}$. Then, MHSA separates these parameters into $M$ heads: $\{\mathbf{W}_l^{\{Q, K, V, O\}, m}\}_{m=1}^M$. As a result, we can denote MHSA as
\begin{align}
&\operatorname{Att}_m(\mathbf{Q}_l, \mathbf{K}_l, \mathbf{V}_l) = \notag \\
&~~~\operatorname{softmax}\left(\frac{\mathbf{Q}_l\mathbf{W}_l^{Q, m}(\mathbf{K}_l\mathbf{W}_l^{K, m})^{T}}{\sqrt{d_k}}\right) \mathbf{V}_l\mathbf{W}_l^{V, m}\mathbf{W}_l^{{O, m}^T}, \notag \\
&\operatorname{MHSA}((\mathbf{Q}_l, \mathbf{K}_l, \mathbf{V}_l)) = \sum^M_{m=1}{\operatorname{Att}_m(\mathbf{Q}_l, \mathbf{K}_l, \mathbf{V}_l)},
\end{align}
where $d_k$ is the dimensionality of the key vectors. The attention heads are computed multiple times in parallel with parameters $\mathbf{W}_l^{Q, m}$, $\mathbf{W}_l^{K, m}$, and $\mathbf{W}_l^{V, m}$. These heads will be transformed linearly through a learnable weight matrix $\mathbf{W}_l^{O, m}$. At last, MHSA derives the final result by concatenating the head outputs.

\begin{algorithm}[t] 
\caption{Process of Apollo} 
\label{alg:apollo}

\begin{algorithmic}[1] 
\REQUIRE the input data $x$, the ground-truth $y$, the stage setting $\{s_t, t\in[1, T], s_t\in [1, S]\}$: a non-decreasing list, indicating the stage of each step.
\FOR{$t=1$ \TO $T$}
    \IF{$t > 1$ \AND $s_t > s_{t-1}$}
        \FOR{$n=1$ \TO $N^{(s)}$}
            \STATE $\theta_n := \text{COPY}\left(\theta_{g_{\text{interpolation}}^{N^{(s_{prev})}:N^{(s)}}(n)}\right)$ 
        \ENDFOR
    \ENDIF
    \STATE $L^{(t)} = \text{LVPS}(N^{(s)}, L)$
    \FOR{$l=1$ \TO $L^{(t)}$}
        \STATE $\Theta(f_{l}) := \text{SHARE}\left(\theta_{g_{\text{interpolation}}^{N^{(s)}:L^{(t)}}(l)}\right)$ 
    \ENDFOR
    \STATE $\mathcal{L} = Loss\left(f^{(L^{(s)})} (x), y\right)$
    \STATE $\{\Delta\theta_i\}^{N^{(t)}} = \mathcal{L}.\operatorname{backward}()$
    \STATE Update all the weights $\{\theta_i\}^{N^{(s)}}$ through $\{\Delta\theta_i\}^{N^{(s)}}$
\ENDFOR 
\ENSURE The trained model $f^{(L)}$ with $\{\theta_i\}^{L}$
\end{algorithmic} 
\end{algorithm}

(2) FFN Layer. The FFN layer applies a non-linear transformation to the output of the self-attention layer by increasing the dimension and then shrinking to the original size.
$\mathbf{X}_l \in \mathbb{R}^{d}$ is the input of the $l$-th layer. The weights are $\mathbf{W}_l^{IN} \in \mathbb{R}^{d\times \alpha d}$ and $\mathbf{W}_l^{OUT} \in \mathbb{R}^{\alpha d\times d}$, where $\alpha$ is an expanding ratio that is often set to 4.
The FFN layer can be formed as
\begin{align}
\operatorname{FFN}(\mathbf{X}_l) = 
\operatorname{GeLU}(\mathbf{X}_l\mathbf{W}_l^{IN})\mathbf{W}_l^{OUT},
\end{align}
where $\operatorname{GeLU}(\cdot)$ is a non-linear activation function~\citep{hendrycks2016gaussian}. The FFN layer plays a crucial role in enhancing the ability to model complex patterns, making it a powerful tool for various natural language processing tasks.

Finally, Applying residual connections to both the MHSA and FFN layers is achieved by incorporating addition and layer normalization operations. These crucial steps serve the purpose of safeguarding the model against collapse and overfitting of the training data. Note that, we neglect biases in the formulation for simplification, which does not affect the generality of the proposed method. Following the above formulation, all weights of the $l$-th layer can be denoted as $\theta_l = \mathbf{W}_l^{\{Q, K, V, O, IN, OUT\}}$.

\begin{figure}[t]
\centering
\includegraphics[width=0.36\textwidth]{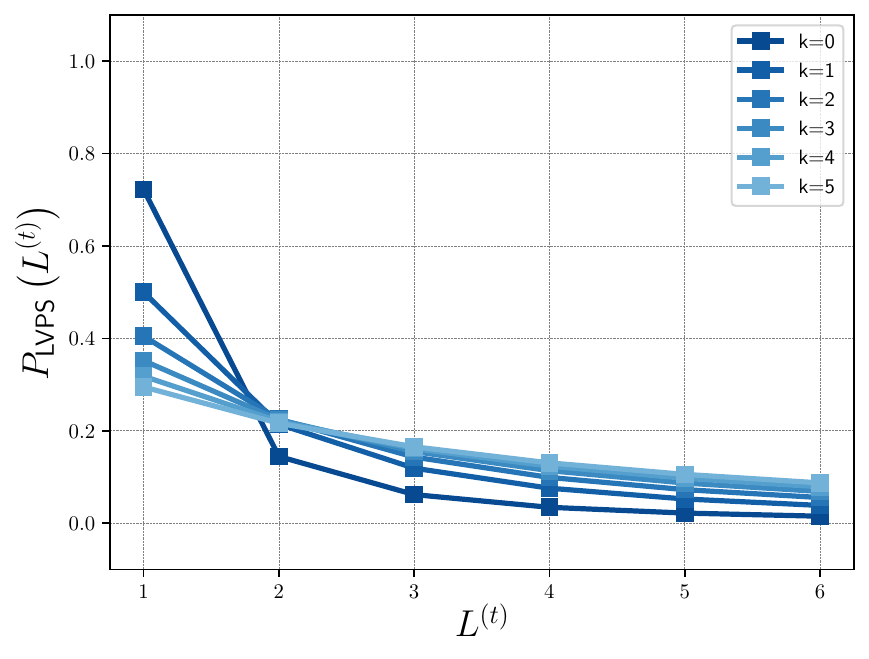}
\caption{A case of choosing hyper-parameters $k$ of LVPS to sample 1-6 layer number.}
\label{fig:lvps}
\end{figure}

\subsection{Efficient Training by Apollo}

Given the potential similarities between high and low layers, the gradual expansion of layers during training can lead to enhanced acceleration, as opposed to the direct approach of training from scratch~\citep{DBLP:conf/icml/GongHLQWL19}. However, as previously discussed, lower layers struggle to effectively capture the intricacies of higher-level features, particularly when utilizing a stacking methodology which can cause training instability. To address this challenge, we introduce Apollo, an innovative approach that facilitates progressive model training by leveraging weight sharing within the training process. This approach enables the accession of high-level functionality prior to layer expansion, thus mitigating the aforementioned issue.

\textbf{Progressive Training.} Specifically in an $L$-layer network, Apollo divides the whole training process into $S$ stages. We use $N^{(s)}, s\in [S]$ to denote the weight number of the $s$-th stage, satisfying
\begin{align}
\label{eq:weight-number}
    N^{(s)} < N^{(s+1)}.
\end{align}
As restricted in Eq.~\eqref{eq:weight-number}, Apollo increases actual weights when stages go on. This progressive training way is efficient for accelerating language models.

\textbf{Preparing Lessons.}  
In order to provide lessons for $N^{(s)}$ weights within the $s$-th stage, thereby facilitating the pre-learning of higher-layer functionalities, Apollo employs a strategic weight-sharing approach. This involves distributing the $N^{(s)}$ weights across $L^{(t)}$ layers, drawn from the range $[N^{(s)}, L]$, thereby establishing a connection to higher-layer elements. The determination of the appropriate $L^{(t)}$ is realized through the utilization of a sampling function denoted as $P(L^{(t)})$ at each training step. The core idea of this sampling function is the prioritization of shallower depths – specifically, $N^{(s)}$ in this context – to strike a balance between computational efficiency and sustained performance. Grounded in this consideration, we introduce a novel approach coined as Low-Value-Prioritized Sampling (LVPS).

\begin{figure}[t]
\centering
\includegraphics[width=0.36\textwidth]{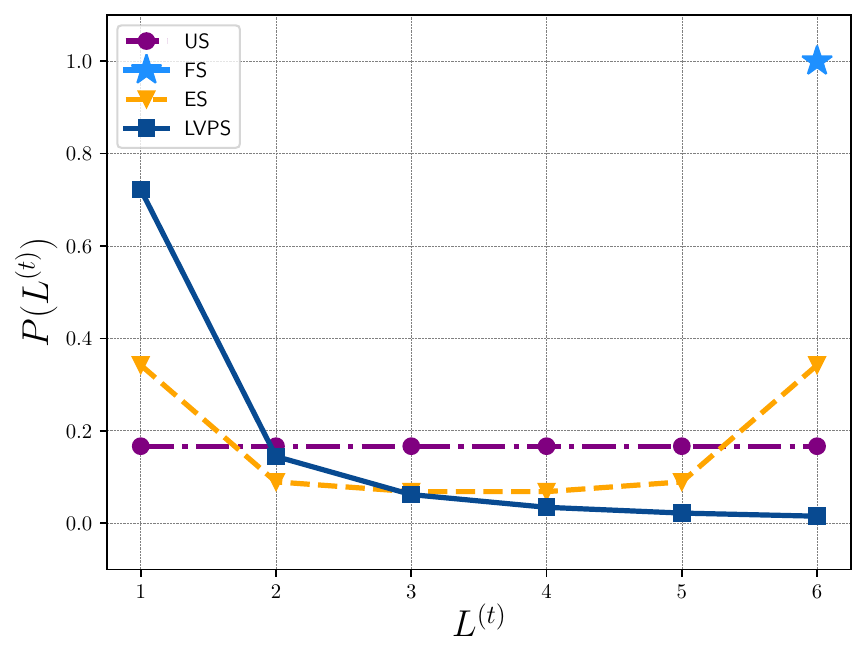}
\caption{Comparison among US, FS, ES, and LVPS to sample 1-6 layer number.}
\label{fig:all-sampling}
\end{figure}

\begin{table}[t]
\centering
\setlength{\tabcolsep}{4pt}
\renewcommand{\arraystretch}{1.2}

\begin{tabular}{@{}c|c@{}}
\toprule
Method & Probability Density Function                                                                                             \\ \midrule
LVPS   & $P_{\text{LVPS}}(L^{(t)}) = \frac{b}{(L^{(t)}+k)^2}$                                                                     \\ \midrule
ES     & $P_{\text{ES}}\left(L^{(t)}\right)=\frac{1}{k} * \left(\frac{1}{L^{(t)} - N^{(s)} - b}+\frac{1}{L + b - L^{(t)}}\right)$ \\ \midrule
US     & $P_{\text{US}}(L^{(t)}) = \frac{1}{L - N^{(s)}}$                                                                         \\ \midrule
FS     & $P_{\text{FS}}(L) = 1$                                                                                                   \\ \bottomrule
\end{tabular}
\caption{An overview of sampling methods: (1) Low-Value-Prioritized Sampling (LVPS), (2) Uniform Sampling (US), (3) Edge Sampling (ES), and (4) Full Sampling (FS). $L^{(t)}$ can only derive values in $[N^{(s)}, L]$. Since the integration of the probability density function is 1, $b$ can be solved when $k$ is determined. In this paper, we set $k=0$ and $k=10$ for LVPS and ES, respectively. FS always samples $L^{(t)}=L$.}
\label{tbl:sampling}
\end{table}

In the development of LVPS, we employ an inverse proportional function as the cumulative distribution function for layer selection with a formulation as
\begin{align}
    F_{\text{LVPS}}(x) = c-\frac{b}{x+k}, c > 0, b > 0, x\in [N^{(s)}, L],
\end{align}
where $b,k$ and $c$ are the hyper-parameters. This function is congruent with our sampling objective, which favors the selection of shallower depths. Then, by setting $F_{\text{LVPS}}\left(N^{(s)}\right)=0$ and $F_{\text{LVPS}}\left(L\right)=1$.
The probability density function $P_{\text{LVPS}}$ can be solved as
\begin{align}
\label{eq:lvps}
P_{\text{LVPS}}(&L^{(t)}) = \begin{cases}
    \frac{b}{(L^{(t)}+k)^2}, & \text{if $L^{(t)} \in [N^{(s)}, L]$},  \\
    0, & \text{otherwise},
\end{cases} \\
\label{eq:lvps-condition}
&\text{w.r.t}~\int P_{\text{LVPS}}(L^{(t)})\d L^{(t)} = 1,
\end{align}
where $b$ and $c$ can be solved in terms of Eq.~\eqref{eq:lvps-condition} as
\begin{align}
    b=\frac{(N^{(s)} + k) * (L + k)}{L - N^{(s)}},~~
    c=\frac{L+k}{L-N^{(s)}}.
\end{align}
Therefore, we only need to adjust $k$ to derive different sampling settings as shown in Fig.~\ref{fig:lvps}. In this paper, we set $k=0$ in every experiment to obtain the least computation complexity.
We employ the notation $\operatorname{LVPS}(\alpha, \beta)$ to signify the process of sampling a value within the interval $[\alpha, \beta]$. $P_{\text{LVPS}}(L^{(t)})$ is the probability function of $\operatorname{LVPS}(N^{(s)}, L)$.

A comparison with other sampling methods can be found in Table~\ref{tbl:sampling} and Fig.~\ref{fig:all-sampling}. In detail, Uniform Sampling (US) samples the layers equally, thereby mitigating the potential layer bias, while Edge  Sampling (ES) tends to sample layers in low and high positions to learn the high-layer information while maintaining efficiency. Alternatively, Full Sampling (FS) always samples the deepest depth, helping early-trained weights adequately acquire the functionality of each layer, which, however, demands significant computational resources due to its selection of the maximum layer number at each step. Among these sampling methods, LVPS can achieve the highest efficiency in progressive training.

\begin{figure}[t]
	\centering
            \label{fig:vit-acc}
            \includegraphics[width=0.43\textwidth]{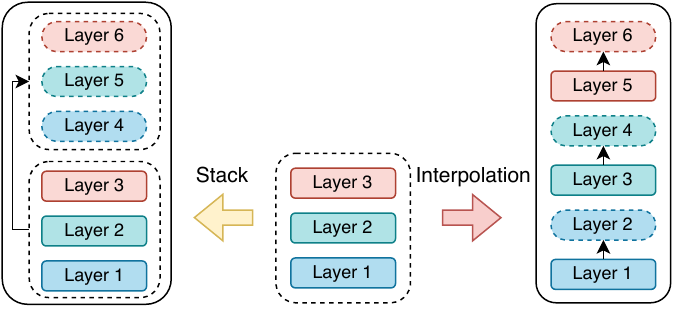}
 
	\caption{A case of expanding 3 layers to 6 layers. The same color denotes the same weight. The stacking method recurrently arranges the layers, e.g., the 1-st layer $\rightarrow$ the 4-th layer. By contrast, the interpolation method arranges the layers in a neighbor, e.g., the 1-st layer $\rightarrow$ the 2-nd layer.}
	\label{fig:expanding}
\end{figure}

\subsection{Stack V.S. Interpolation}

In the training process of Apollo, it is important to select a feasible method for expanding the layer for sharing in each step and initializing weights of the next stage. As shown in Fig.~\ref{fig:expanding}, the common-used methods are stacking and interpolating layers. Given a target to expand $L_1$ to $L_2$, the stacking method can be formed as
\begin{align}
\label{eq:stack}
g^{L_1:L_2}_{\text{stack}}(l_2) = l_2 \bmod{L_1},
\end{align}
where $l_2 \in [L_2]$ is the index of $L_2$.
The stacking method is usually adopted in language models, e.g., BERT~\citep{DBLP:conf/icml/GongHLQWL19}. By contrast, the interpolation method is often used in the computer vision field, e.g., ResNet~\citep{DBLP:conf/iclr/ChangMHTB18}, and can be formulated as
\begin{align}
\label{eq:inter}
g^{L_1:L_2}_{\text{interpolation}}(l_2) = \lfloor\frac{l_2\ast L_1}{L_2}\rceil,
\end{align}
where $\ast$ means dot production, and $\lfloor \cdot \rceil$ denotes the rounding operation.
Although the two expanding methods have both shown good performance in their fields, there is still a lack of analysis on the comparison between them, especially in the applicability of language models. In this paper, we investigate the influence of the stability and performance of these methods. We defer an analysis experiment to the experiment section. Here, we would like to give the conclusion: (1) There is only a small performance gap between them; (2) interpolation can reach better stability than the stacking method. As a result, we adopt interpolation instead of the stacking method as the expanding method of the proposed Apollo training for language models.

\section{Experiment}
\label{sec:exp}
In this section, we conduct a series of experiments to validate the performance of our proposed method.

\paragraph{Common setting:} We implement a series of experiments on BERT and GPT for validation. We use AdamW as the optimizer with a learning rate of $10^{-4}$ and weight decay of $10^{-2}$ in all the experiments. We chose the training batch sizes of 768 and 512 for BERT~\citep{DBLP:conf/naacl/DevlinCLT19} and GPT~\citep{radford2019language} models, respectively. We use the Scratch model, StackBERT, bert2BERT, and LiGO as baselines. Layer numbers of Apollo are [1, 3, 6, 12] and change at epoch [2, 4, 10]. LiGO is warmly trained for 100 steps as claimed in the original paper~\citep{wanglearning}. The training dataset is a concatenation of English Wikipedia and Toronto Book Corpus~\cite{DBLP:conf/iccv/ZhuKZSUTF15}. The structures of the BERT and GPT models are in Table~\ref{tbl:structurebertgpt} of the Appendix.

\begin{figure}[t]
\centering
\includegraphics[width=0.38\textwidth]{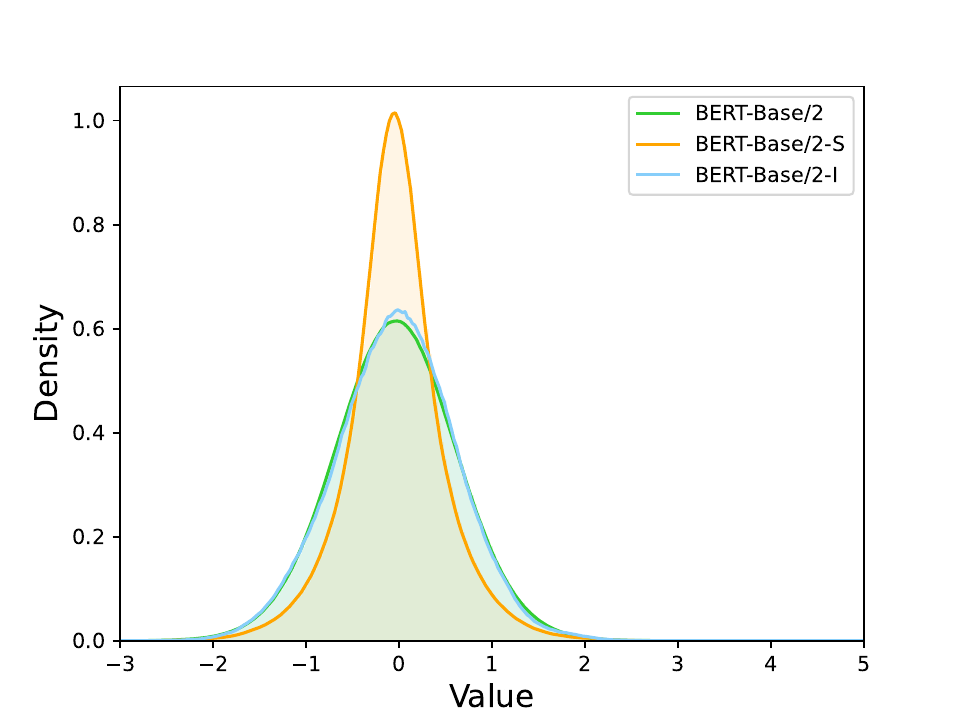}

\caption{Distribution of output activations. BERT-Base/2 is half of BERT-Base. BERT-Base/2-S and BERT-Base/2-I denote to stack and interpolate BERT-Base/2 to BERT-Base, respectively. After stacking BERT-Base/2, the distribution of output activations changes a lot, while the interpolation method keeps the distribution well.}
\label{fig:act-distri}
\end{figure}

\begin{table}[t]

\centering
\setlength{\tabcolsep}{3.8pt}
\renewcommand{\arraystretch}{1.2}

\scalebox{0.98}{
\begin{tabular}{@{}lcccc@{}}
\toprule
Model         & Trainable & Layer & Loss  & Gradient (1e-3)      \\ \midrule
BERT-Base Ra.        & -         & 12    & 10.56 & 11.45$\pm$26.46  \\ \midrule
BERT-Base/2   & -         & 6     & 1.82  & 1.66$\pm$4.71    \\
BERT-Base/2-S & \ding{55} & 12    & 7.32  & 55.28$\pm$241.52 \\
BERT-Base/2-I & \ding{51} & 12    & 3.97  & 44.01$\pm$225.76 \\ \bottomrule
\end{tabular}
}
\caption{Results of the analysis of expanding methods. BERT-Base Ra. means a randomly initialized BERT-Base. When expanding BERT-Base/2 from 6 to 12, the loss and gradients of the stacking method rise sharply, which causes failure in further training.
By contrast, the interpolation method achieves smaller gradients and loss and is thus trainable later. However, both of the two methods cause higher gradients than a random model, indicating an unstable state.
}
\label{tbl:basic-expand-size}
\end{table}

\begin{table}[t]

\centering
\setlength{\tabcolsep}{4.5pt}
\renewcommand{\arraystretch}{1.2}

\scalebox{0.98}{
\begin{tabular}{@{}lcccc@{}}
\toprule
Model                     & Acc. Ratio              & Layer & Loss & Gradient (1e-3)   \\ \midrule
\multirow{2}{*}{Apollo-S} & \multirow{2}{*}{39.7\%} & 6     & 1.82 & 1.65$\pm$3.72 \\
                          &                         & 12    & 1.76 & 1.42$\pm$3.50 \\ \midrule
\multirow{2}{*}{Apollo-I} & \multirow{2}{*}{41.6\%} & 6     & 1.82 & 1.59$\pm$3.33 \\
                          &                         & 12    & 1.76 & 1.39$\pm$3.01 \\ \bottomrule
\end{tabular}
}
\caption{Results of expanding methods for Apollo. ``S'' and ``I'' mean the stacking and the interpolation methods, respectively. Apollo can decrease the gradient values loss after expanding layers since learning the functionality of a 12-layer network. Moreover, the interpolation method can derive a comparably better acceleration ratio in terms of FLOPs and smaller gradient values.
}
\label{tbl:apollo-expand-size}
\end{table}

\begin{figure*}[t]
\centering
\subfigure[On BERT-Base (FLOPs).]{
        \label{fig:bert-loss-flops}
        \includegraphics[width=0.31\textwidth]{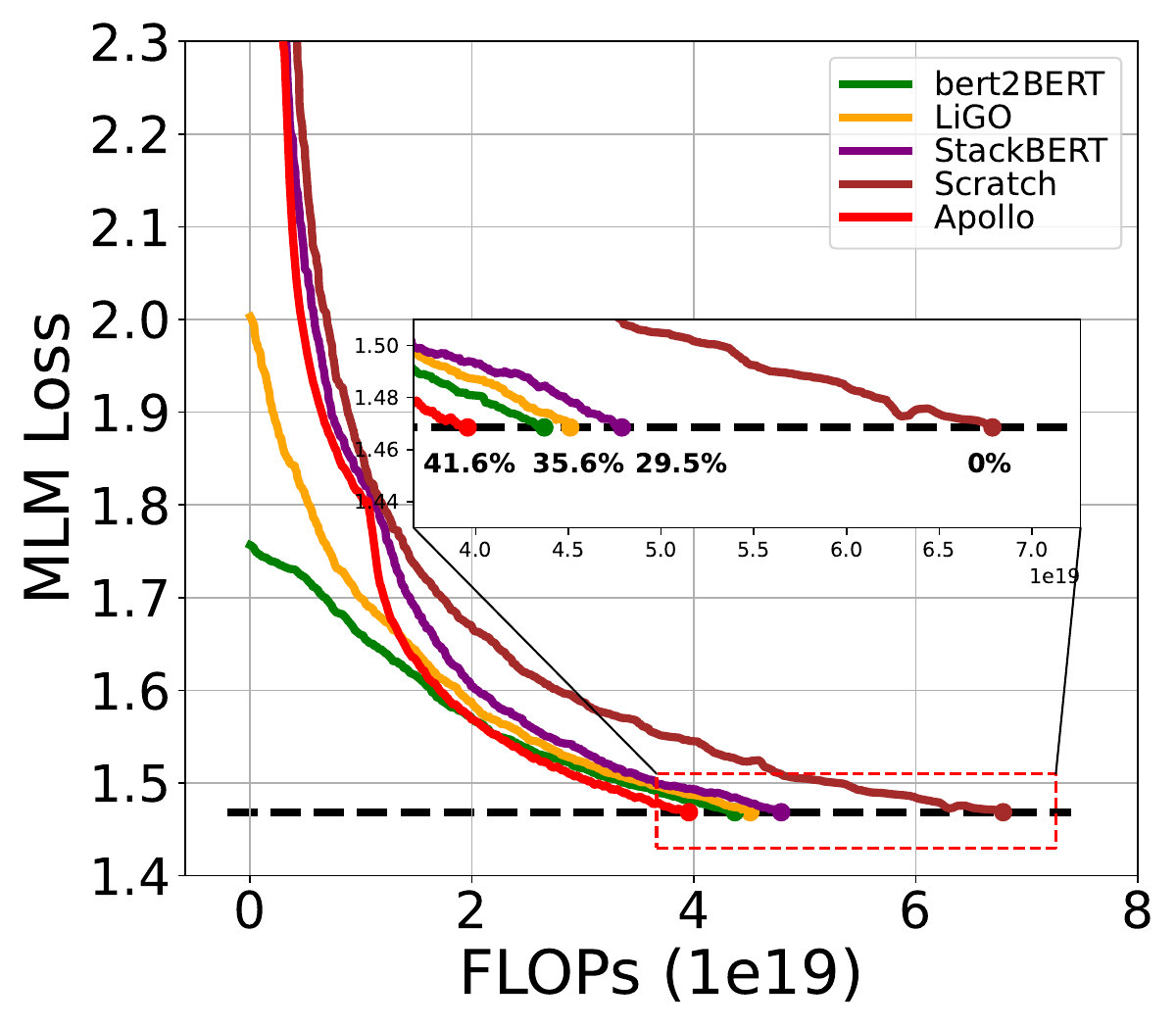}

}
\subfigure[On BERT-Base (Wall Time).]{
        \label{fig:bert-loss-walltime}
        \includegraphics[width=0.31\textwidth]{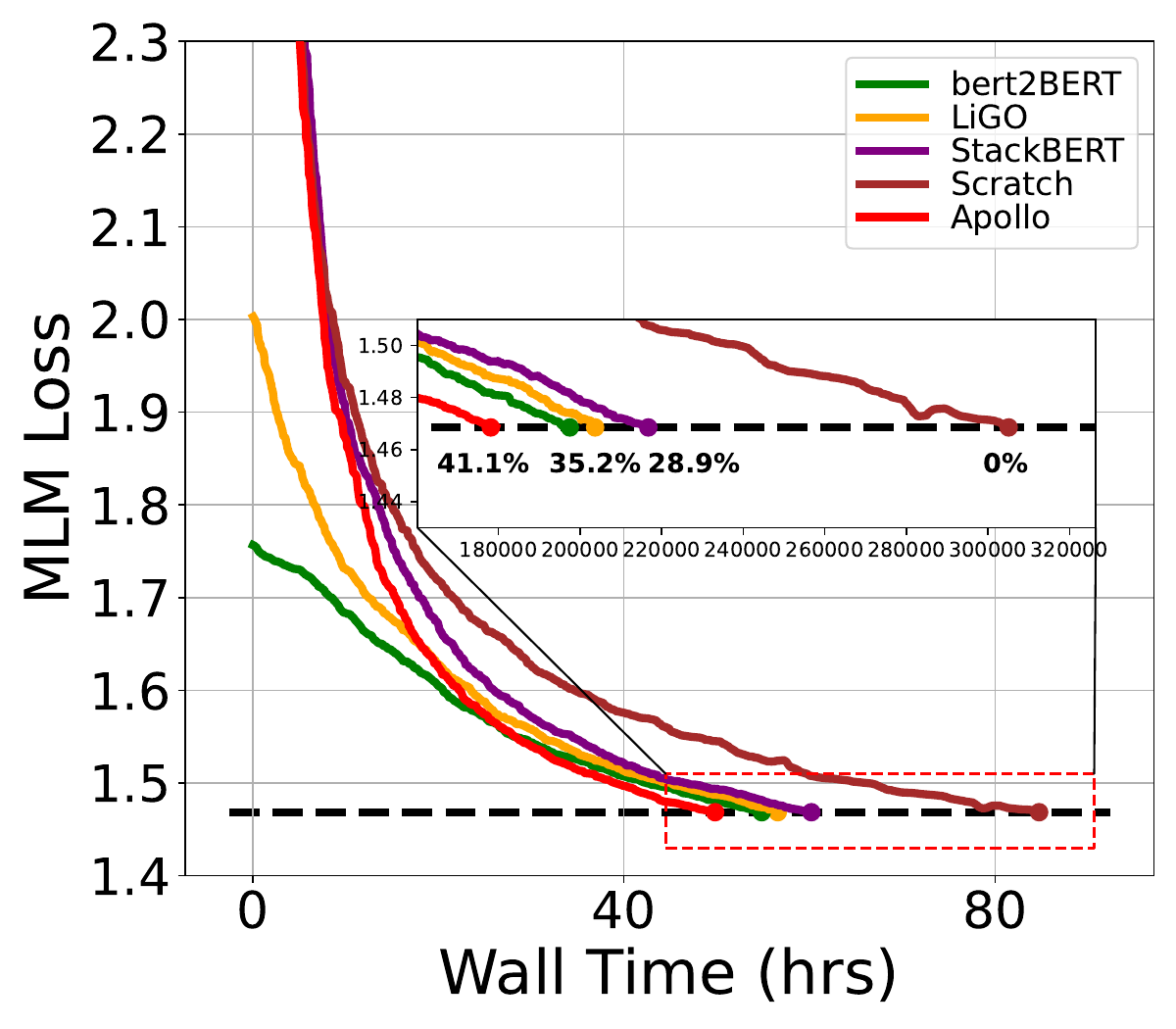}

}
\subfigure[On GPT-Base (FLOPs).]{
        \label{fig:gpt-loss-flops}
        \includegraphics[width=0.31\textwidth]{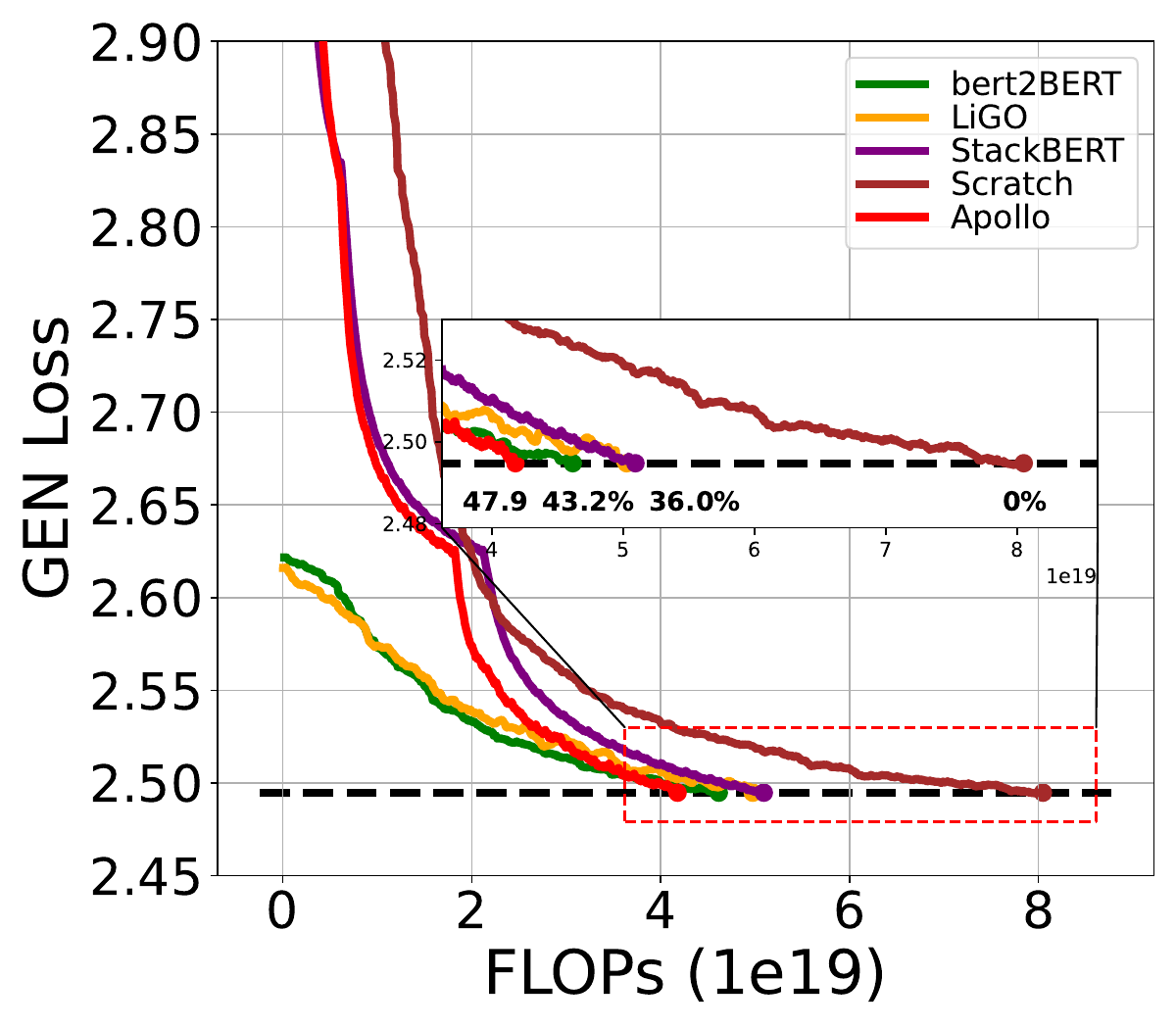}

}
\caption{Results of BERT-Base and GPT-Base. Apollo achieves the highest acceleration on BERT-Base and GPT-Base in terms of FLOPs at 41.6\% and 47.9\%, respectively. In addition, Apollo can keep the best training efficiency on wall time for BERT-Base at 41.1\%. Apollo surpasses methods (i.e., bert2BERT and LiGO) relying on pretrained models in all cases.}
\label{fig:base-result}
\end{figure*}

\subsection{Experiment on Expanding Method}

\begin{figure}[t]
\centering
\includegraphics[width=0.36\textwidth]{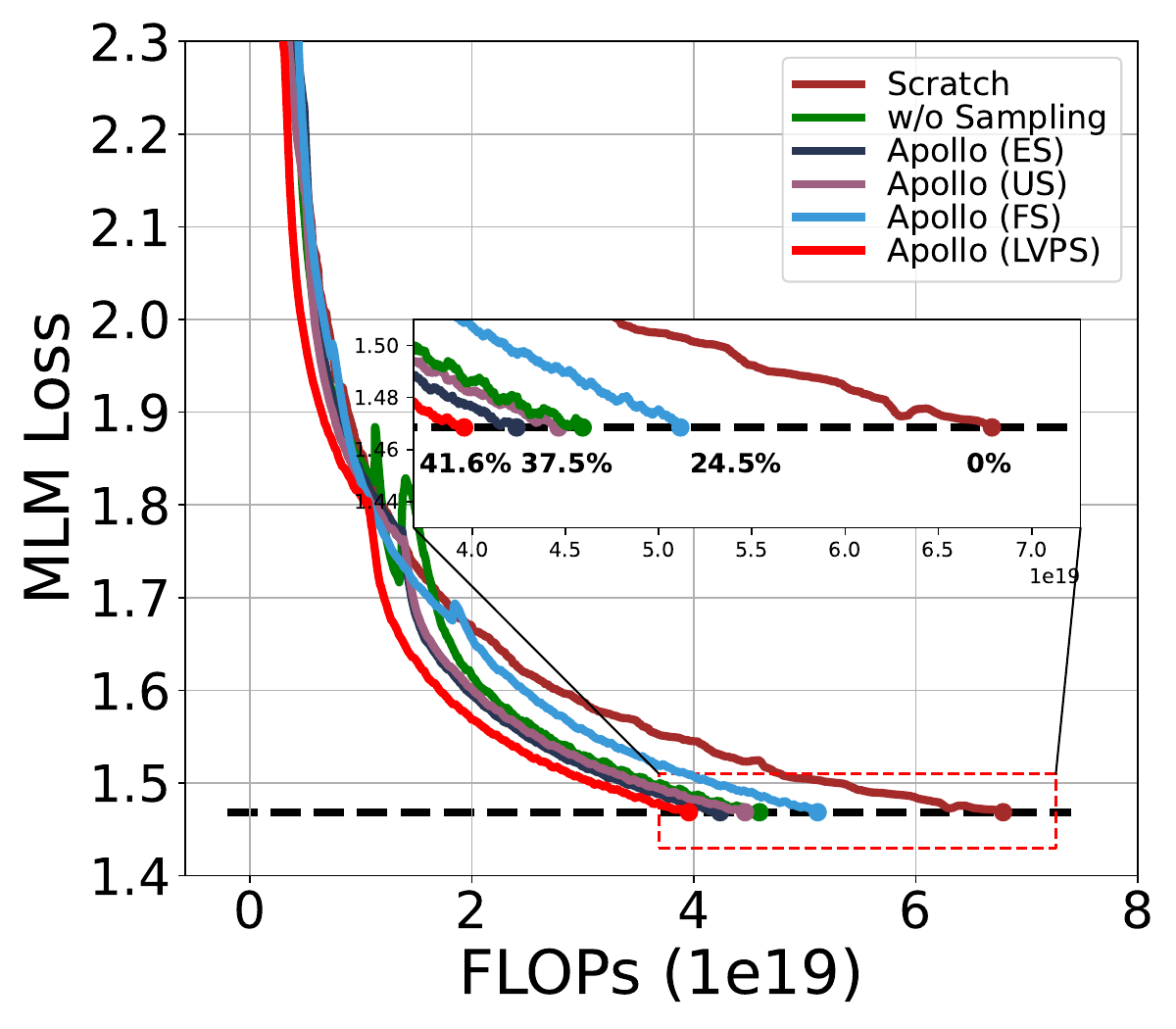}

\caption{Results of the analysis on sampling methods. LVPS achieves the highest acceleration ratio at 41.6\%, surpassing ES and FS at 4.1\% and +17.1\%, respectively. Most sampling methods are faster than Apollo w/o Sampling. FS performs the lowest acceleration by always sampling the deepest depth, which is resource-consuming.}
\label{fig:exp-sampling}
\end{figure}

We implement the experiment to show the influence of stacking and interpolating layers. We use BERT as the backbone. We train a 6-layer BERT model called BERT-Base/2 which is the half of BERT-Base for 10 epochs with a 768 batch size.
Then, we expand the trained BERT-Base/2 into the 12-layer BERT-Base through the stacking (Eq.~\eqref{eq:stack}) and interpolating (Eq.~\eqref{eq:inter}) methods. We use 500 data samples for validation. Moreover, we apply Apollo to train BERT-Base with stacking and interpolation methods.

Illustrated in Fig.\ref{fig:act-distri}, directly stacking BERT-Base/2 exhibits a notable alteration in the distribution of output activations, whereas the interpolation method preserves this distribution. This distinction results in interpolation yielding smaller losses and gradients in comparison to the stacking technique. This advantage contributes to the enhanced trainability of BERT-Base-I, as evidenced in Table\ref{tbl:basic-expand-size}. Additionally, in Table~\ref{tbl:apollo-expand-size}, where expanding BERT-Base/2 increases both loss and gradients, Apollo showcases a reverse trend. Here, layer expansion within Apollo leads to a reduction in the loss and gradient for the lessons on the functionality of 12 layers, ensuring robust stability in progressive learning. Notably, interpolation further diminishes gradient values and improves the acceleration ratio, measured in terms of FLOPs. Therefore, we adopt interpolation as the preferred expansion method for Apollo considering these findings.

\begin{table*}[ht]
\centering
\setlength{\tabcolsep}{2pt}
\renewcommand{\arraystretch}{1.2}

\scalebox{0.96}{
\begin{tabular}{@{}lccccccccccccc@{}}
\toprule
Model     & \begin{tabular}[c]{@{}c@{}}Saving\\ (FLOPs)\end{tabular} & \begin{tabular}[c]{@{}c@{}}Saving\\ (Wall Time)\end{tabular} & \begin{tabular}[c]{@{}c@{}}SQuADv1.1\\ (F1)\end{tabular} & \begin{tabular}[c]{@{}c@{}}SQuADv2.0\\ (F1)\end{tabular} & \begin{tabular}[c]{@{}c@{}}SST-2\\ (Acc)\end{tabular} & \begin{tabular}[c]{@{}c@{}}MNLI\\ (Acc)\end{tabular} & \begin{tabular}[c]{@{}c@{}}MRPC\\ (Acc)\end{tabular} & \begin{tabular}[c]{@{}c@{}}COLA\\ (Mcc)\end{tabular} & \begin{tabular}[c]{@{}c@{}}QNLI\\ (Acc)\end{tabular} & \begin{tabular}[c]{@{}c@{}}STS-B\\ (Acc)\end{tabular} & \begin{tabular}[c]{@{}c@{}}QQP\\ (Acc)\end{tabular} & \begin{tabular}[c]{@{}c@{}}GLUE\\ Avg.\end{tabular} & \begin{tabular}[c]{@{}c@{}}SQuAD\\ Avg.\end{tabular} \\ \midrule
Scratch   & -                                                        & -                                                           & 89.05                                                    & 77.49                                                    & 92.04                                                 & 84.05                                                & 87.65                                                & 56.95                                                & 91.39                                                & 89.16                                                 & 91.17                                               & 84.63                                               & 83.27                                                \\ \midrule
\multicolumn{14}{l}{Training from the Pretrained Model: BERT-Small$\rightarrow$BERT-Base}                                                                                                                                                                                                                                                                                                                                                                                                                                                                                                                                                                                                                                                                               \\ \midrule
bert2BERT & 35.6\%                                                   & 35.2\%                                                      & 90.02                                                    & 78.99                                                    & 92.89                                                 & 84.92                                                & 86.91                                                & 60.32                                                & 91.81                                                & 88.11                                                 & 90.72                                               & 85.10                                               & 84.50                                                \\
LiGO      & 33.5\%                                                   & 33.2\%                                                      & 90.09                                                    & 78.34                                                    & 92.75                                                 & 84.99                                                & 87.44                                                & 61.10                                                & 91.33                                                & 87.94                                                 & 90.42                                               & 85.14                                               & 84.22                                                \\ \midrule
\multicolumn{14}{l}{Progressive Training form Scratch}                                                                                                                                                                                                                                                                                                                                                                                                                                                                                                                                                                                                                                                                                                                  \\ \midrule
StackBERT & 29.5\%                                                   & 28.9\%                                                      & 89.82                                                    & 78.21                                                    & 92.94                                                 & 84.63                                                & 87.65                                                & 61.61                                                & 90.95                                                & 87.13                                                 & 90.20                                               & 85.01                                               & 84.01                                                \\
Apollo    & \textbf{41.6\%}                                          & \textbf{41.1\%}                                             & 89.87                                                    & 78.42                                                    & 92.28                                                 & 84.81                                                & 87.06                                                & 60.57                                                & 91.43                                                & 88.27                                                 & 90.69                                               & 85.02                                               & 84.15                                                \\ \bottomrule
\end{tabular}

}
\caption{Experiments on downstream tasks of BERT-Base on GLUE~\citep{DBLP:conf/iclr/WangSMHLB19}, SQuADv1.1~\citep{DBLP:conf/emnlp/RajpurkarZLL16}, and SQuADv2.0~\citep{DBLP:conf/acl/RajpurkarJL18} dataset. The terms ``Training from the Pretrained Model'' and ``Progressive Training from Scratch'' denote the pretraining type of the methods in the table. Compared with baselines, Apollo can achieve the highest FLOPs saving under similar downstream performance, even better than training from the pretrained model.}
\label{tbl:bertds}
\end{table*}

\subsection{Experiment on Sampling Method}

We construct the experiment on BERT-Base to investigate the influence of the sampling methods including ES, US, FS, and LVPS, with a comparison to the w/o sampling case.

Results are shown in Fig.~\ref{fig:exp-sampling}. The LVPS sampling method leads with the highest acceleration ratio, attaining an impressive 41.6\%. Both ES and LVPS attach the top two positions for acceleration, underscoring the utility of sampling lower layers. Significantly, LVPS outperforms Apollo w/o sampling by a substantial margin of +9.3\%, affirming the pronounced benefits of extracting high-layer functionalities through sampling. US performing normally may indicate that sampling middle layers is not really helpful. FS demonstrates the least acceleration due to its consistent sampling of the maximum layers, incurring a substantial FLOP cost. In summation, LVPS strategically harnesses the ability to capture high-layer functionalities while simultaneously minimizing FLOP usage by primarily sampling lower layers.

\begin{figure}[t]
\centering
\includegraphics[width=0.36\textwidth]{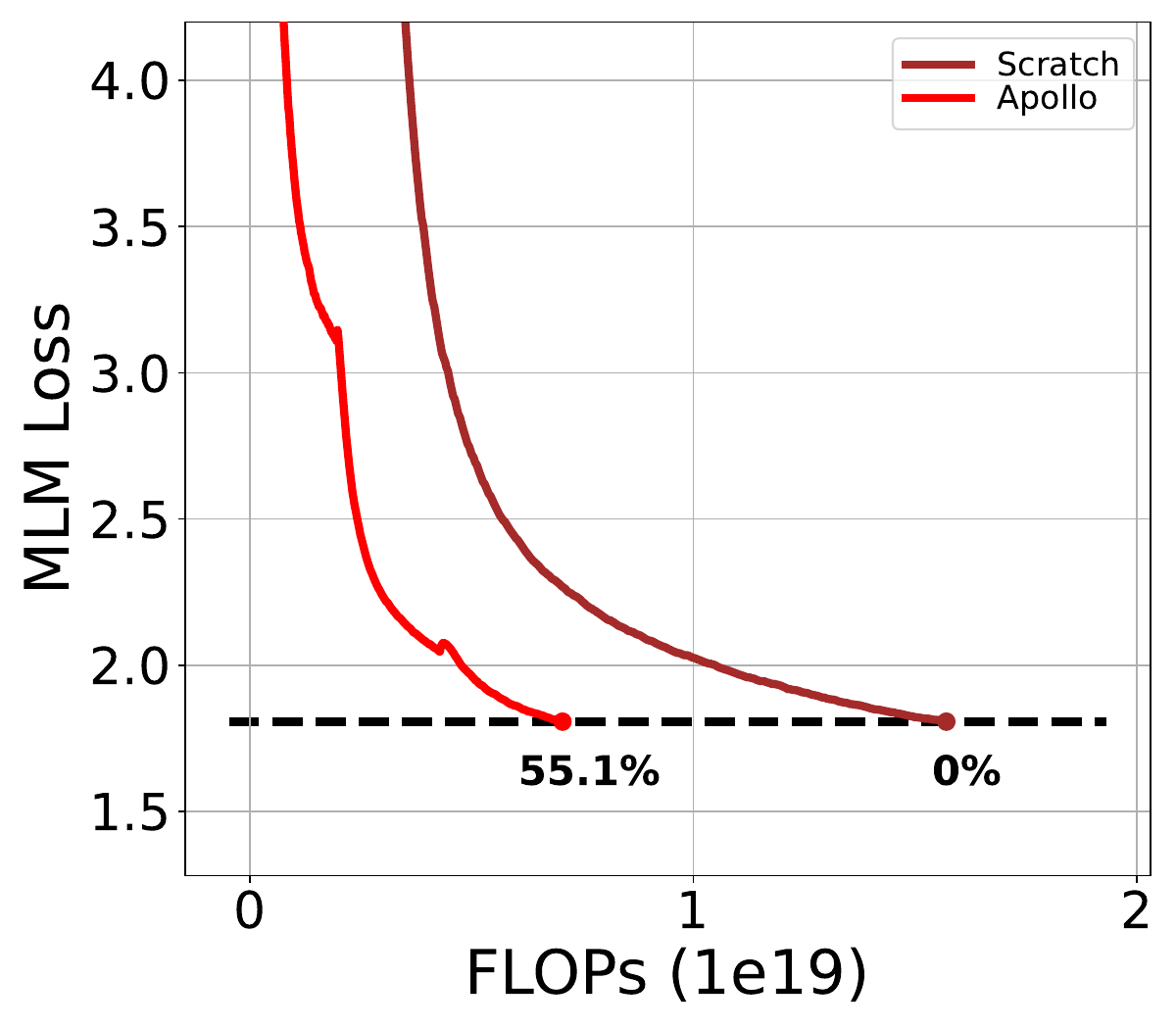}
\caption{Results of BERT-Large.}
\label{fig:bert-large-loss}
\end{figure}

\subsection{Experiment on BERT}
We construct this experiment to show the training efficiency of BERT. We train Apollo and StackBERT from scratch, while training bert2BERT and LiGO from BERT-Small to BERT-Base for 40 epochs. Additionally, we also implement short training on BERT-Large for 5 epochs.

As shown in Fig.~\ref{fig:base-result}, Apollo achieves the highest acceleration rate of 41.6\% in FLOPs saving, which outperforms bert2BERT by an additional +6.0\%. Through training from a pretrained BERT-Small, bert2BERT and LiGO exhibit superior performance compared to the progressive training approach StackBERT, resulting in acceleration ratios of 29.5\% with corresponding improvements of +6.1\% and +4.0\%, respectively. However, the convergence speed of both bert2BERT and LiGO is found to be slower in comparison to Apollo, which shows the accelerated ability of the Apollo method through the ``lessons'' of higher layers.
To substantiate the effectiveness of Apollo, comprehensive evaluations were conducted using the SQuAD and GLUE benchmark datasets, as illustrated in Table~\ref{tbl:bertds}. For a larger BERT-Large in Fig.~\ref{fig:bert-large-loss}, Apollo also shows a significant acceleration ratio at 55.1\%. The results clearly demonstrate the ability of Apollo to exhibit proficient transfer learning capabilities, coupled with faster convergence rates, thus making it promising for practical applications.

\subsection{Experiment on GPT}
We also implement the experiment to validate the performance of GPT. We train Apollo, StackBERT and Scratch from a random initialization, and bert2BERT and LiGO from GPT-Small. The training epoch is 35.

As shown in Fig.~\ref{fig:gpt-loss-flops}, we construct a comparison among the Scratch model, StackBERT, bert2BERT, LiGO, and Apollo. Notably, our proposed Apollo method exhibits a remarkable 47.9\% acceleration ratio. Despite the structural disparities between GPT and BERT, encompassing distinct features such as the mask method, and the position of layer normalization, Apollo consistently maintains the highest performance across the evaluated models. Specifically, Apollo shows a significant advancement over StackBERT, achieving an improvement of +11.9\% for the prepared lessons before expanding. Furthermore, compared to bert2BERT and LiGO, Apollo attains a considerable acceleration advantage with +6.8\% and +9.9\% higher performance, respectively. These findings highlight the substantial acceleration capabilities of the Apollo method, affirming its effectiveness for broader applications.

\section{Conclusion}
\label{sec:conclusion}

Training language models imposes a substantial demand on computational resources. Compared to training from pretrained models, progressive training from scratch offers a universal and flexible solution to accelerate training, as it does not require a pretrained model. However, previous progressive methods suffered from inefficient layer expansion, leading to suboptimal training efficiency.
In light of these challenges, we propose Apollo, a novel approach that imparts lessons for weights in the early stage, achieving a more natural and efficient expansion. Additionally, we conduct a thorough analysis of the influence between stacking and interpolating methods for expanding model depth, advocating the use of interpolation for improved stability in progressive training.
Experimental results consistently demonstrate that Apollo achieves remarkable acceleration across various language models, even surpassing methods that rely on pretrained models. This implies the significant effectiveness of Apollo in enhancing training efficiency. In the future, we expect that Apollo will contribute to the realization of green AI by significantly reducing the cost of training Transformers.

\section{Acknowledgments}
This work was partially supported by the National Natural Science Foundation of China (No. 62276002), a key program of fundamental research from Shenzhen Science and Technology Innovation Commission (No. JCYJ20200109113403826), the Major Key Project of PCL (No. 2022ZD0115301), and an Open Research Project of Zhejiang Lab (NO.2022RC0AB04).

\bibliography{aaai24}

\begin{thebibliography}{40}
\providecommand{\natexlab}[1]{#1}

\bibitem[{Bolya et~al.(2022)Bolya, Fu, Dai, Zhang, Feichtenhofer, and Hoffman}]{DBLP:journals/corr/abs-2210-09461}
Bolya, D.; Fu, C.; Dai, X.; Zhang, P.; Feichtenhofer, C.; and Hoffman, J. 2022.
\newblock Token Merging: Your ViT But Faster.
\newblock \emph{CoRR}, abs/2210.09461.

\bibitem[{Cao et~al.(2021)Cao, Wang, Chen, Jiang, Zhang, Tian, and Wang}]{DBLP:journals/corr/abs-2105-05537}
Cao, H.; Wang, Y.; Chen, J.; Jiang, D.; Zhang, X.; Tian, Q.; and Wang, M. 2021.
\newblock Swin-Unet: Unet-like Pure Transformer for Medical Image Segmentation.
\newblock \emph{CoRR}, abs/2105.05537.

\bibitem[{Chang et~al.(2018)Chang, Meng, Haber, Tung, and Begert}]{DBLP:conf/iclr/ChangMHTB18}
Chang, B.; Meng, L.; Haber, E.; Tung, F.; and Begert, D. 2018.
\newblock Multi-level Residual Networks from Dynamical Systems View.
\newblock In \emph{{ICLR} (Poster)}. OpenReview.net.

\bibitem[{Chen et~al.(2022)Chen, Yin, Shang, Jiang, Qin, Wang, Wang, Chen, Liu, and Liu}]{DBLP:conf/acl/ChenYS0QWWCL022}
Chen, C.; Yin, Y.; Shang, L.; Jiang, X.; Qin, Y.; Wang, F.; Wang, Z.; Chen, X.; Liu, Z.; and Liu, Q. 2022.
\newblock bert2BERT: Towards Reusable Pretrained Language Models.
\newblock In \emph{{ACL} {(1)}}, 2134--2148. Association for Computational Linguistics.

\bibitem[{Chen, Goodfellow, and Shlens(2016)}]{DBLP:journals/corr/ChenGS15}
Chen, T.; Goodfellow, I.~J.; and Shlens, J. 2016.
\newblock Net2Net: Accelerating Learning via Knowledge Transfer.
\newblock In \emph{{ICLR}}.

\bibitem[{Dehghani et~al.(2019)Dehghani, Gouws, Vinyals, Uszkoreit, and Kaiser}]{DBLP:conf/iclr/DehghaniGVUK19}
Dehghani, M.; Gouws, S.; Vinyals, O.; Uszkoreit, J.; and Kaiser, L. 2019.
\newblock Universal Transformers.
\newblock In \emph{{ICLR} (Poster)}. OpenReview.net.

\bibitem[{Devlin et~al.(2019)Devlin, Chang, Lee, and Toutanova}]{DBLP:conf/naacl/DevlinCLT19}
Devlin, J.; Chang, M.; Lee, K.; and Toutanova, K. 2019.
\newblock {BERT:} Pre-training of Deep Bidirectional Transformers for Language Understanding.
\newblock In \emph{{NAACL-HLT} {(1)}}, 4171--4186. Association for Computational Linguistics.

\bibitem[{Dosovitskiy et~al.(2021)Dosovitskiy, Beyer, Kolesnikov, Weissenborn, Zhai, Unterthiner, Dehghani, Minderer, Heigold, Gelly, Uszkoreit, and Houlsby}]{DBLP:conf/iclr/DosovitskiyB0WZ21}
Dosovitskiy, A.; Beyer, L.; Kolesnikov, A.; Weissenborn, D.; Zhai, X.; Unterthiner, T.; Dehghani, M.; Minderer, M.; Heigold, G.; Gelly, S.; Uszkoreit, J.; and Houlsby, N. 2021.
\newblock An Image is Worth 16x16 Words: Transformers for Image Recognition at Scale.
\newblock In \emph{{ICLR}}. OpenReview.net.

\bibitem[{Gong et~al.(2019)Gong, He, Li, Qin, Wang, and Liu}]{DBLP:conf/icml/GongHLQWL19}
Gong, L.; He, D.; Li, Z.; Qin, T.; Wang, L.; and Liu, T. 2019.
\newblock Efficient Training of {BERT} by Progressively Stacking.
\newblock In \emph{{ICML}}, volume~97 of \emph{Proceedings of Machine Learning Research}, 2337--2346. {PMLR}.

\bibitem[{Gu et~al.(2021)Gu, Liu, Yu, Li, Chen, and Han}]{DBLP:conf/naacl/GuLYLCH21}
Gu, X.; Liu, L.; Yu, H.; Li, J.; Chen, C.; and Han, J. 2021.
\newblock On the Transformer Growth for Progressive {BERT} Training.
\newblock In \emph{{NAACL-HLT}}, 5174--5180. Association for Computational Linguistics.

\bibitem[{Hendrycks and Gimpel(2016)}]{hendrycks2016gaussian}
Hendrycks, D.; and Gimpel, K. 2016.
\newblock Gaussian error linear units (gelus).
\newblock \emph{arXiv preprint arXiv:1606.08415}.

\bibitem[{Kamalakara et~al.(2022)Kamalakara, Locatelli, Venkitesh, Ba, Gal, and Gomez}]{DBLP:journals/corr/abs-2209-13569}
Kamalakara, S.~R.; Locatelli, A.; Venkitesh, B.; Ba, J.; Gal, Y.; and Gomez, A.~N. 2022.
\newblock Exploring Low Rank Training of Deep Neural Networks.
\newblock \emph{CoRR}, abs/2209.13569.

\bibitem[{Kingma and Ba(2015)}]{DBLP:journals/corr/KingmaB14}
Kingma, D.~P.; and Ba, J. 2015.
\newblock Adam: {A} Method for Stochastic Optimization.
\newblock In \emph{{ICLR} (Poster)}.

\bibitem[{Lan et~al.(2020)Lan, Chen, Goodman, Gimpel, Sharma, and Soricut}]{DBLP:conf/iclr/LanCGGSS20}
Lan, Z.; Chen, M.; Goodman, S.; Gimpel, K.; Sharma, P.; and Soricut, R. 2020.
\newblock {ALBERT:} {A} Lite {BERT} for Self-supervised Learning of Language Representations.
\newblock In \emph{{ICLR}}. OpenReview.net.

\bibitem[{Li et~al.(2022)Li, Zhuang, Wang, Liang, Chang, and Yang}]{DBLP:conf/cvpr/LiZWLCY22}
Li, C.; Zhuang, B.; Wang, G.; Liang, X.; Chang, X.; and Yang, Y. 2022.
\newblock Automated Progressive Learning for Efficient Training of Vision Transformers.
\newblock In \emph{{CVPR}}, 12476--12486. {IEEE}.

\bibitem[{Li et~al.(2020)Li, Pan, Chen, Ding, Zhao, and Xu}]{DBLP:journals/corr/abs-2009-10580}
Li, N.; Pan, Y.; Chen, Y.; Ding, Z.; Zhao, D.; and Xu, Z. 2020.
\newblock Heuristic Rank Selection with Progressively Searching Tensor Ring Network.
\newblock \emph{CoRR}, abs/2009.10580.

\bibitem[{Pan et~al.(2019)Pan, Xu, Wang, Ye, Wang, Bai, and Xu}]{DBLP:conf/aaai/PanXWYWBX19}
Pan, Y.; Xu, J.; Wang, M.; Ye, J.; Wang, F.; Bai, K.; and Xu, Z. 2019.
\newblock Compressing Recurrent Neural Networks with Tensor Ring for Action Recognition.
\newblock In \emph{{AAAI}}, 4683--4690. {AAAI} Press.

\bibitem[{Pan et~al.(2023)Pan, Yuan, Yin, Xu, Shang, Jiang, and Liu}]{DBLP:journals/corr/abs-2310-10699}
Pan, Y.; Yuan, Y.; Yin, Y.; Xu, Z.; Shang, L.; Jiang, X.; and Liu, Q. 2023.
\newblock Reusing Pretrained Models by Multi-linear Operators for Efficient Training.
\newblock \emph{CoRR}, abs/2310.10699.

\bibitem[{Qin et~al.(2022)Qin, Lin, Yi, Zhang, Han, Zhang, Su, Liu, Li, Sun, and Zhou}]{DBLP:conf/naacl/QinLYZ0ZSLLS022}
Qin, Y.; Lin, Y.; Yi, J.; Zhang, J.; Han, X.; Zhang, Z.; Su, Y.; Liu, Z.; Li, P.; Sun, M.; and Zhou, J. 2022.
\newblock Knowledge Inheritance for Pre-trained Language Models.
\newblock In \emph{{NAACL-HLT}}, 3921--3937. Association for Computational Linguistics.

\bibitem[{Radford et~al.(2019)Radford, Wu, Child, Luan, Amodei, Sutskever et~al.}]{radford2019language}
Radford, A.; Wu, J.; Child, R.; Luan, D.; Amodei, D.; Sutskever, I.; et~al. 2019.
\newblock Language models are unsupervised multitask learners.
\newblock \emph{OpenAI blog}, 1(8): 9.

\bibitem[{Rajpurkar, Jia, and Liang(2018)}]{DBLP:conf/acl/RajpurkarJL18}
Rajpurkar, P.; Jia, R.; and Liang, P. 2018.
\newblock Know What You Don't Know: Unanswerable Questions for SQuAD.
\newblock In \emph{{ACL} {(2)}}, 784--789. Association for Computational Linguistics.

\bibitem[{Rajpurkar et~al.(2016)Rajpurkar, Zhang, Lopyrev, and Liang}]{DBLP:conf/emnlp/RajpurkarZLL16}
Rajpurkar, P.; Zhang, J.; Lopyrev, K.; and Liang, P. 2016.
\newblock SQuAD: 100, 000+ Questions for Machine Comprehension of Text.
\newblock In \emph{{EMNLP}}, 2383--2392. The Association for Computational Linguistics.

\bibitem[{Rogers, Kovaleva, and Rumshisky(2020)}]{DBLP:journals/tacl/RogersKR20}
Rogers, A.; Kovaleva, O.; and Rumshisky, A. 2020.
\newblock A Primer in BERTology: What We Know About How {BERT} Works.
\newblock \emph{Trans. Assoc. Comput. Linguistics}, 8: 842--866.

\bibitem[{Schwartz et~al.(2019)Schwartz, Dodge, Smith, and Etzioni}]{DBLP:journals/corr/abs-1907-10597}
Schwartz, R.; Dodge, J.; Smith, N.~A.; and Etzioni, O. 2019.
\newblock Green {AI}.
\newblock \emph{CoRR}, abs/1907.10597.

\bibitem[{Shen et~al.(2022)Shen, Walsh, Keutzer, Dodge, Peters, and Beltagy}]{DBLP:conf/icml/ShenWKDPB22}
Shen, S.; Walsh, P.; Keutzer, K.; Dodge, J.; Peters, M.~E.; and Beltagy, I. 2022.
\newblock Staged Training for Transformer Language Models.
\newblock In \emph{{ICML}}, volume 162 of \emph{Proceedings of Machine Learning Research}, 19893--19908. {PMLR}.

\bibitem[{Shoeybi et~al.(2019)Shoeybi, Patwary, Puri, LeGresley, Casper, and Catanzaro}]{DBLP:journals/corr/abs-1909-08053}
Shoeybi, M.; Patwary, M.; Puri, R.; LeGresley, P.; Casper, J.; and Catanzaro, B. 2019.
\newblock Megatron-LM: Training Multi-Billion Parameter Language Models Using Model Parallelism.
\newblock \emph{CoRR}, abs/1909.08053.

\bibitem[{Vaswani et~al.(2017)Vaswani, Shazeer, Parmar, Uszkoreit, Jones, Gomez, Kaiser, and Polosukhin}]{DBLP:conf/nips/VaswaniSPUJGKP17}
Vaswani, A.; Shazeer, N.; Parmar, N.; Uszkoreit, J.; Jones, L.; Gomez, A.~N.; Kaiser, L.; and Polosukhin, I. 2017.
\newblock Attention is All you Need.
\newblock In \emph{{NIPS}}, 5998--6008.

\bibitem[{Wang et~al.(2019{\natexlab{a}})Wang, Singh, Michael, Hill, Levy, and Bowman}]{DBLP:conf/iclr/WangSMHLB19}
Wang, A.; Singh, A.; Michael, J.; Hill, F.; Levy, O.; and Bowman, S.~R. 2019{\natexlab{a}}.
\newblock {GLUE:} {A} Multi-Task Benchmark and Analysis Platform for Natural Language Understanding.
\newblock In \emph{{ICLR} (Poster)}. OpenReview.net.

\bibitem[{Wang et~al.(2019{\natexlab{b}})Wang, Li, Wu, Chandra, and Liu}]{DBLP:journals/corr/abs-1910-03103}
Wang, D.; Li, M.; Wu, L.; Chandra, V.; and Liu, Q. 2019{\natexlab{b}}.
\newblock Energy-Aware Neural Architecture Optimization with Fast Splitting Steepest Descent.
\newblock \emph{CoRR}, abs/1910.03103.

\bibitem[{Wang et~al.(2023{\natexlab{a}})Wang, Pan, Yang, Li, Xu, and Cichocki}]{DBLP:journals/corr/abs-2302-09019}
Wang, M.; Pan, Y.; Yang, X.; Li, G.; Xu, Z.; and Cichocki, A. 2023{\natexlab{a}}.
\newblock Tensor Networks Meet Neural Networks: {A} Survey and Future Perspectives.
\newblock \emph{CoRR}, abs/2302.09019.

\bibitem[{Wang et~al.(2020)Wang, Su, Luo, Pan, Zheng, and Xu}]{DBLP:conf/iconip/WangSL0ZX20}
Wang, M.; Su, Z.; Luo, X.; Pan, Y.; Zheng, S.; and Xu, Z. 2020.
\newblock Concatenated Tensor Networks for Deep Multi-Task Learning.
\newblock In \emph{{ICONIP} {(5)}}, volume 1333 of \emph{Communications in Computer and Information Science}, 517--525. Springer.

\bibitem[{Wang et~al.(2023{\natexlab{b}})Wang, Panda, Hennigen, Greengard, Karlinsky, Feris, Cox, Wang, and Kim}]{wanglearning}
Wang, P.; Panda, R.; Hennigen, L.~T.; Greengard, P.; Karlinsky, L.; Feris, R.; Cox, D.~D.; Wang, Z.; and Kim, Y. 2023{\natexlab{b}}.
\newblock Learning to Grow Pretrained Models for Efficient Transformer Training.
\newblock In \emph{{ICLR}}. OpenReview.net.

\bibitem[{Wu et~al.(2020{\natexlab{a}})Wu, Liu, Stone, and Liu}]{DBLP:conf/nips/WuLS020}
Wu, L.; Liu, B.; Stone, P.; and Liu, Q. 2020{\natexlab{a}}.
\newblock Firefly Neural Architecture Descent: a General Approach for Growing Neural Networks.
\newblock In \emph{NeurIPS}.

\bibitem[{Wu, Wang, and Liu(2019)}]{DBLP:conf/nips/WuW019}
Wu, L.; Wang, D.; and Liu, Q. 2019.
\newblock Splitting Steepest Descent for Growing Neural Architectures.
\newblock In \emph{NeurIPS}, 10655--10665.

\bibitem[{Wu et~al.(2020{\natexlab{b}})Wu, Ye, Lei, Lee, and Liu}]{DBLP:journals/corr/abs-2003-10392}
Wu, L.; Ye, M.; Lei, Q.; Lee, J.~D.; and Liu, Q. 2020{\natexlab{b}}.
\newblock Steepest Descent Neural Architecture Optimization: Escaping Local Optimum with Signed Neural Splitting.
\newblock \emph{CoRR}, abs/2003.10392.

\bibitem[{Wu et~al.(2021)Wu, Xing, Li, Ke, He, and Liu}]{DBLP:conf/iclr/0001XLKHL21}
Wu, Q.; Xing, C.; Li, Y.; Ke, G.; He, D.; and Liu, T. 2021.
\newblock Taking Notes on the Fly Helps Language Pre-Training.
\newblock In \emph{{ICLR}}. OpenReview.net.

\bibitem[{Yang et~al.(2020)Yang, Wang, Yang, Li, He, and Zhang}]{DBLP:journals/corr/abs-2011-13635}
Yang, C.; Wang, S.; Yang, C.; Li, Y.; He, R.; and Zhang, J. 2020.
\newblock Progressively Stacking 2.0: {A} Multi-stage Layerwise Training Method for {BERT} Training Speedup.
\newblock \emph{CoRR}, abs/2011.13635.

\bibitem[{Yang et~al.(2021)Yang, Hou, Song, Liu, and Zhou}]{DBLP:journals/corr/abs-2110-03848}
Yang, S.; Hou, L.; Song, X.; Liu, Q.; and Zhou, D. 2021.
\newblock Speeding up Deep Model Training by Sharing Weights and Then Unsharing.
\newblock \emph{CoRR}, abs/2110.03848.

\bibitem[{Zhang and He(2020)}]{DBLP:conf/nips/ZhangH20}
Zhang, M.; and He, Y. 2020.
\newblock Accelerating Training of Transformer-Based Language Models with Progressive Layer Dropping.
\newblock In \emph{NeurIPS}.

\bibitem[{Zhu et~al.(2015)Zhu, Kiros, Zemel, Salakhutdinov, Urtasun, Torralba, and Fidler}]{DBLP:conf/iccv/ZhuKZSUTF15}
Zhu, Y.; Kiros, R.; Zemel, R.~S.; Salakhutdinov, R.; Urtasun, R.; Torralba, A.; and Fidler, S. 2015.
\newblock Aligning Books and Movies: Towards Story-Like Visual Explanations by Watching Movies and Reading Books.
\newblock In \emph{{ICCV}}, 19--27. {IEEE} Computer Society.

\end{thebibliography}

\clearpage
\appendix

\section{Appendix}

\subsection{Hyper-parameter Choice of Edge Sampling (ES)}
\begin{figure}[h]
\centering
\includegraphics[width=0.38\textwidth]{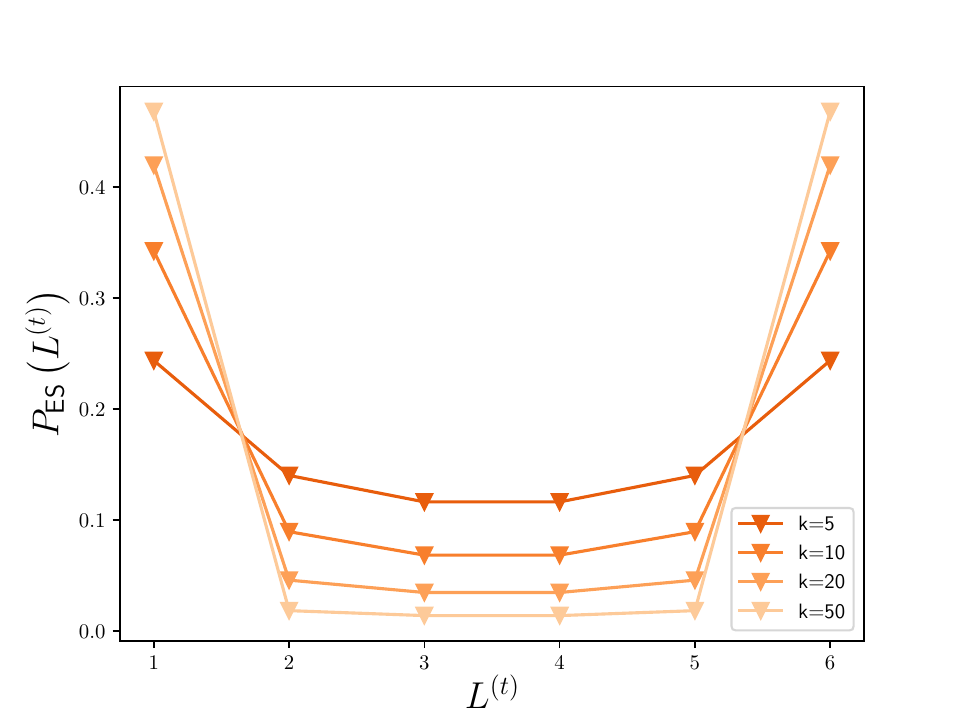}
\caption{A case of choosing hyper-parameters $k$ of ES to sample 1-6 layer number.}
\label{fig:hyper-es}
\end{figure}

Edge Sampling (ES) mainly samples low and high depths, while seldom sampling the middle part. As formulated in Table~\ref{tbl:sampling}, ES can be determined by selecting $k$ as $b$ can be solved when $k$ is chosen. We illustrate the influence of the hyper-parameter $k$ of ES in Fig.~\ref{fig:hyper-es}. We choose $k=10$ for ES in all our experiments, since it offers a clearer differentiation between edges and middle regions, without exhibiting excessive bias towards edges. Hence, it presents a balanced and practical option for comparison.

\subsection{Structures of Language Models}
We show structures of the used language models including the BERT and GPT models in Table~\ref{tbl:structurebertgpt}.

\begin{table}[ht]
\centering
\setlength{\tabcolsep}{4pt}
\renewcommand{\arraystretch}{1.0}
\scalebox{0.7}{
\begin{tabular}{@{}c|cccccc@{}}
\toprule
Config      & BERT-Small & BERT-Base & BERT-Large & GPT-Small & GPT-Base \\ \midrule
\# layers   & 12         & 12        & 24         & 12        & 12       \\
\# hidden   & 512        & 768       & 1024       & 512       & 768      \\
\# heads    & 8          & 12        & 16         & 8         & 12       \\
\# vocab    & 30522      & 30522     & 30522      & 50257     & 50257    \\
seq. length & 512        & 512       & 512        & 1024      & 1024     \\ \bottomrule
\end{tabular}
}
\caption{The structures of BERT and GPT.}
\label{tbl:structurebertgpt}
\end{table}

\end{document}